\pdfoutput=1

\documentclass[11pt]{article}

\usepackage[]{EMNLP2022}
\usepackage{float}
\usepackage{hyperref}
\usepackage{times}
\usepackage{latexsym}
\usepackage{amsmath}
\usepackage[T1]{fontenc}

\usepackage[utf8]{inputenc}
\usepackage{stfloats}
\usepackage{microtype}

\usepackage{siunitx}
\usepackage{graphicx} 
\usepackage{float} 
\usepackage{subfigure} 
\usepackage{multirow}
\usepackage{booktabs}
\usepackage{makecell}
\title{Multi-View Reasoning: \\
Consistent Contrastive Learning for Math Word Problem}

\author{Wenqi Zhang$^{1}$,
  Yongliang Shen$^{1}$,
  Yanna Ma$^{2}$,
  Xiaoxia Cheng$^{1}$, \\
  {\bf Zeqi Tan$^{1}$,}
  {\bf Qingpeng Nong$^{3}$,}
  {\bf Weiming Lu$^{1}$\footnotemark[2]}\\
  $^1$College of Computer Science and Technology, Zhejiang University \\
  $^2$University of Shanghai for Science and Technology\\
  $^3$Zhongxing Telecommunication Equipment Corporationy\\
  \texttt{\{zhangwenqi, luwm\}@zju.edu.cn} } 
\begin{document}
\maketitle

\renewcommand{\thefootnote}{\fnsymbol{footnote}} 
\footnotetext[2]{Corresponding author.}  
\renewcommand{\thefootnote}{\arabic{footnote}}

\begin{abstract}
Math word problem solver requires both precise relation reasoning about quantities in the text and reliable generation for the diverse equation. Current sequence-to-tree or relation extraction methods regard this only from a fixed view, struggling to simultaneously handle complex semantics and diverse equations. However, human solving naturally involves two consistent reasoning views: top-down and bottom-up, just as math equations also can be expressed in multiple equivalent forms: pre-order and post-order. We propose a multi-view consistent contrastive learning for a more complete semantics-to-equation mapping. The entire process is decoupled into two independent but consistent views: top-down decomposition and bottom-up construction, and the two reasoning views are aligned in multi-granularity for consistency, enhancing global generation and precise reasoning. Experiments on multiple datasets across two languages show our approach significantly outperforms the existing baselines, especially on complex problems \footnote{Our source code and data are 
open sourced at \url{https://github.com/zwq2018/Multi-view-Consistency-for-MWP}}. We also show after consistent alignment, multi-view can absorb the merits of both views and generate more diverse results consistent with the mathematical laws.
\end{abstract}
\section{Introduction}
Math word problem (MWP) is a very significant and challenging task with a wide range of applications in both natural language processing and general artificial intelligence \citep{bobrow1964natural}. The MWP is to predict the mathematical equation and the final answer based on a natural language description of the scenario and a math problem.
It requires mathematical reasoning over the text \citep{mukherjee2008review}, which is very challenging for conventional methods \citep{patel-etal-2021-nlp}.\\
\begin{figure}[t] 
\centering 
\includegraphics[width=0.49\textwidth]{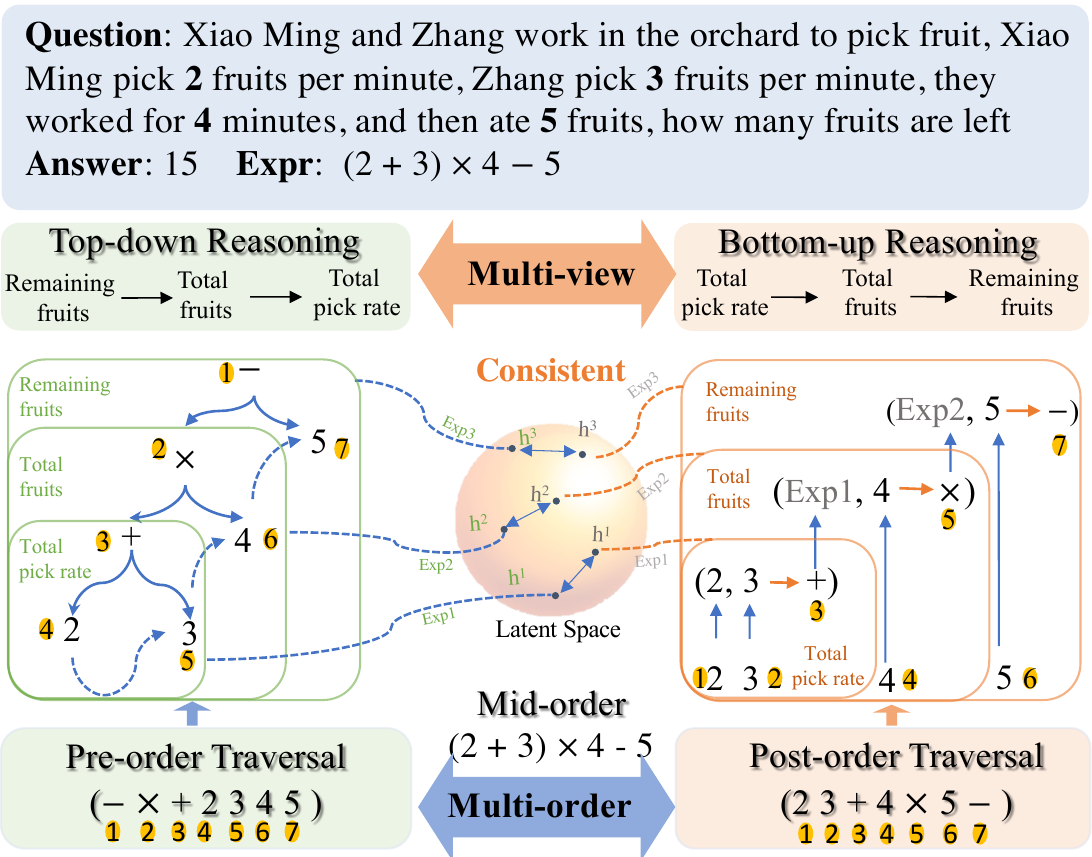} 
\caption{Human solving has multiple reasoning views, and math equation also can be expressed in multi-order. Pre-order traversal can be seen as a top-down reasoning view. Post-order traversal corresponds exactly to the bottom-up reasoning view. Consistent contrastive learning aligns two views in the same latent space.} 
\label{Fig.f1} 
\end{figure}
\indent MWP tasks have attracted a great deal of research attention. In the early days, MWP was treated as a sequence-to-sequence (seq2seq) translation task, translating human language into mathematical language \citep{wang2017deep, wang2019template}. Then, \citet{xie2019goal, zhang2020graph,faldu2021towards} proposed that tree or graph structure was more suitable for MWP. Those generation methods (Seq2Tree and Graph2Tree) further improved generation capabilities through a specific structure. Although very flexible in generating complex equation combinations, the fixed structure decoder also limits its fine-grained mapping. Recently, \citet{cao2021bottom, jie2022learning} introduced an iterative relation extraction approach, providing a new solving view for MWP. It performs well at capturing local relations, but lacks global generation capabilities, especially for complex mathematical problems.\\
\indent From the seq2seq translation to the seq2tree generation and relation extraction, those are essentially seeking a suitable solving view for MWP. However, MWP is more challenging than that as it requires both precise relation reasoning about quantities and reliable generation for diverse equation combinations. Both are necessary for mathematical reasoning. Existing methods all consider the MWP from a single view and thus bring certain limitations.\\
\indent We argue that multiple views are required to comprehensively
solve the MWP. As shown in Figure \ref{Fig.f1}, the process of human solving inherently involves multiple reasoning views, i.e., top-down decomposition ($remaining\ fruits \! \stackrel{-}\rightarrow \! total\ fruits \! \stackrel{\times}\rightarrow \! pick\ rate\stackrel{+}\rightarrow$), and bottom-up construction ($\stackrel{+}\rightarrow pick\ rate\! \stackrel{\times}\rightarrow \! total\ fruits \! \stackrel{-}\rightarrow \! remaining\ fruits$). Two reasoning views are reversed in the process but consistent in results. Meanwhile, mathematical equation can be expressed in multi-order traversal, i.e., pre-order ($-,\times,+,2,3,4,5$) and post-order ($2,3,+,4,\times,5,-$). Two sequences are quite dissimilar in form but equivalent in logic. Two order traversal equation corresponds exactly to the two reasoning processes, i.e. , the pre-order equation is a top-down reasoning view, while the post-order can be seen as a bottom-up reasoning view.\\
\indent Inspired by this, we design multi-view reasoning using multi-order traversal. The MWP solving is decoupled into two independent but consistent views: top-down reasoning using pre-order traversal to decompose problem from global to local and a bottom-up process following post-order traversal for relation construction from local to global. Pre-order and post-order traversals should be equivalent in math just as top-down decomposition and bottom-up construction should be consistent. In Figure \ref{Fig.f1}, we add multi-granularity contrastive learning to align the intermediate expressions generated by two views in the same latent space. Through consistent alignment, two views constrain each other and jointly learn a accurate and complete representation for math reasoning.\\
\indent Besides, math operator must conform to mathematical laws (e.g., commutative law). We devise a knowledge-enhanced augmentation to incorporate mathematical rules into the learning process, promoting multi-view reasoning more consistent with mathematical rules.\\
\indent Our contributions are threefold:

\vspace{-\topsep}
\begin{itemize}
  \setlength{\parskip}{2pt}
  \setlength{\itemsep}{0pt plus 1pt}
\item We treat multi-order traversal as a multi-view reasoning process, which contains a top-down decomposition using pre-order traversal and a down-up construction following post-order. Both views are necessary for MWP.
\item We introduce consistent contrastive learning to align two views reasoning processes, fusing flexible global generation and accurate semantics-to-equation mapping. We also design an augmentation process for rules injection and understanding.
\item Extensive experiments on multiple standard datasets show our method signiﬁcantly outperforms existing baselines. Our method can also generate equivalent but non-annotated math equations, demonstrating reliable reasoning ability behind our multi-view framework.
\end{itemize}
\vspace{-\topsep}




\section{Related Work}
Reliable reasoning is a necessary capability to move towards general-purpose AI. How to achieve human-like reasoning has been extensively researched in areas such as natural language processing, reinforcement learning, and robotics \citep{ijcai2021p328, Zhang2021LearningTN, ijcai2022p654}. In particular, mathematical reasoning is an important manifestation of intelligence. Automatically solving mathematical problems has been studied for a long time, from rule-based methods \citep{fletcher1985understanding, bakman2007robust,yuhui2010frame} with hand-crafted features and templates-based methods \citep{kushman2014learning, 10.1162/tacl_a_00012} to deep learning methods \citep{wang2017deep, ling-etal-2017-program} with the encoder-decoder framework. The introduction of Transformer \citep{vaswani2017attention} and pre-trained language models \citep{devlin-etal-2019-bert,liu2019roberta} greatly improves the performance of MWPs. From the perspective of proxy tasks, we divide the recent works into three categories: seq2seq-based translation, seq2structure generation, and iterative relation extraction.\\
\indent \textbf{Seq2seq-based translation} MWPs are treated as a translation task, translating human language into mathematical language \citep{ijcai2021-485}. \citet{wang2017deep} proposed a large-scale dataset Math23K and used the vanilla seq2seq method \citep{chiang-chen-2019-semantically}. \citet{li-etal-2019-modeling} introduced a group attention mechanism to enhance seq2seq method performance. \citet{huang-etal-2018-neural} used reinforcement learning to optimize translation task. \citet{huang-etal-2017-learning} incorporated semantic-parsing methods to solve MWPs. Although seq2seq-based methods have made great progress in the field, the performance of these methods is still unsatisfying, since the generation of mathematical equations requires relation reasoning over quantities than natural language.\\
\indent \textbf{Seq2structure-based generation} \citet{liu-etal-2019-tree, xie2019goal} introduced tree-structured decoder to generate mathematical expressions. This explicit tree-based design rapidly dominated the MWPs community. Other researchers have begun to explore reasonable structures for encoder. \citet{li-etal-2020-graph-tree,zhang2020graph,HGEN} used graph neural networks to extract effective logical information from the natural language problem. \citet{ijcai2021-485} adopted the teacher model using contrast learning to improve the encoder. Several researchers have attempted to extract multi-level features from the problems using the hierarchical encoder \citep{lin2021hms} and pre-trained model \citep{yu-etal-2021-improving}. Many auxiliary tasks are used to enhance the symbolic reasoning ability \citep{qin-etal-2021-neural}. \citet{wu2020knowledge,wu-etal-2021-math} tried to introduce mathematical knowledge to solve the difficult mathematical reasoning. These structured generation approaches show strong generation capabilities towards complex mathematical reasoning tasks.\\
\indent \textbf{Iterative relation extraction} Recently, some researchers have borrowed ideas from the field of information extraction \citep{shenyongliang}, and have designed iterative relation extraction frameworks for predicting math relations between two numeric tokens. \citet{kim-etal-2020-point} designed an expression-pointer transformer model to predict expression fragmentation. \citet{cao2021bottom} introduced a DAG structure to extract numerical token relation from bottom to top. \citet{jie2022learning} further treated the MWP task as an iterative relation extraction task, achieving impressive performance. These works provide a new perspective to tackle MWP from a local relation construction view, improving the fine-grained relation reasoning between quantities.\\
\indent The above proxy tasks are designed from different solving views. The seq2seq is a left-to-right consecutive view, while seq2tree is a tree view, and the relation extraction method emphasizes a local relation view. Unlike these single-view methods, our approach employs multiple consistent reasoning views to address the challenges of MWP.
\section{Approach}
\subsection{Overview} \label{section3.1}
The MWP is to predict the equation $Y$ and the answer based on a problem description $T\ \text{=}\ \{w_{1},w_{2} \cdots w_{n}\}$ containing $n$ words and $m$ quantity words $Q\ \text{=}\ \{q_1,q_2,\cdots,q_{m}\}$. The equation $Y$ is a sequence of constant words (e.g., 3.14), mathematical operator $op\ \text{=}\ \{+, -, \times, \div,\cdots\}$ and quantity words from $Q$. Solving MWP is to find the optimal mapping $T\rightarrow \hat{Y}$, allowing predicted $\hat{Y}$ to derive the correct answer. Existing methods learn this mapping from a single view, e.g., seq2tree generation and iterative relation extraction. Our consistent contrastive learning approach solves this by reasoning from multiple views. Both top-down and bottom-up view are necessary for a complete semantics-to-equation mapping. 
\subsection{Multi-View using Multi-Order}
\begin{figure*}[htbp] 
\centering 
\includegraphics[width=1\textwidth]{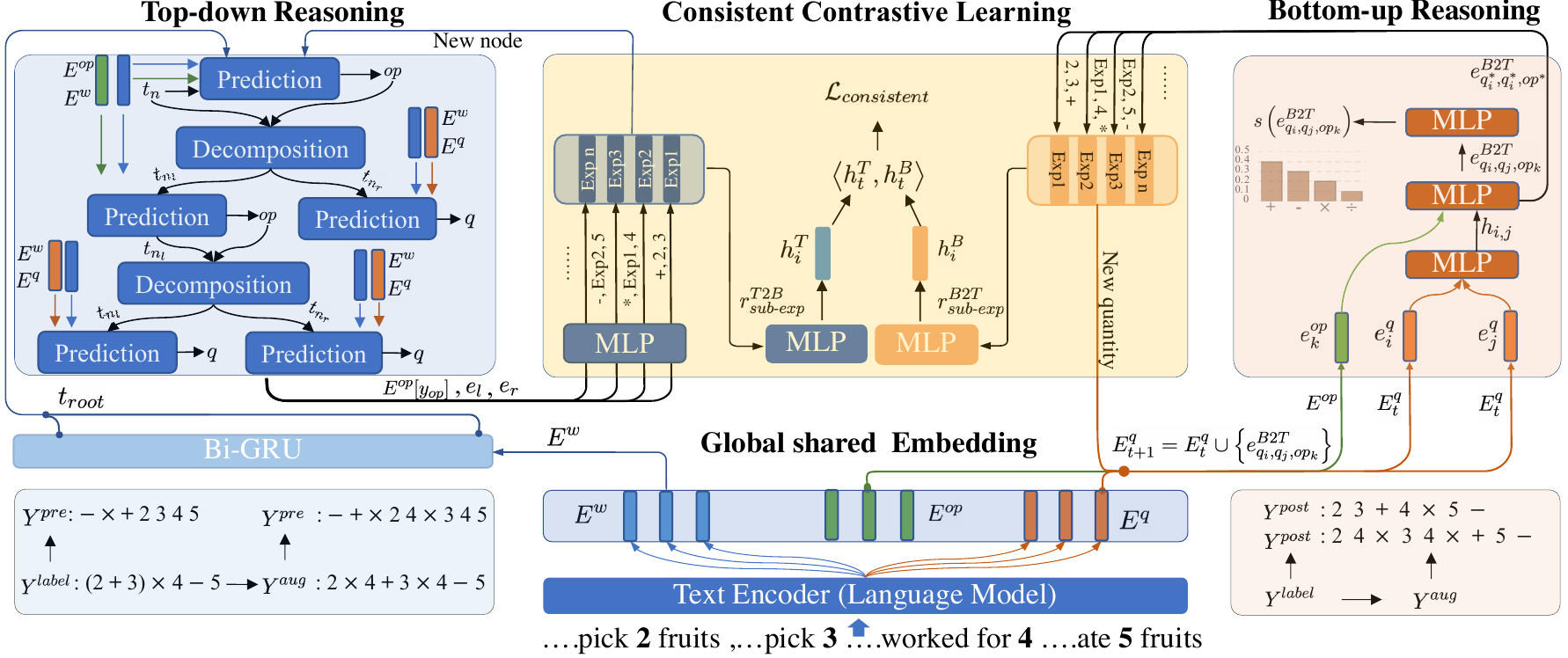} 
\caption{We align multi-view reasoning through consistent contrastive learning. Firstly, we obtain three global shared embeddings ($E^{op},E^{q},E^{w}$) from the text encoder. Then, top-down process constantly decomposes global goals, while bottom-up process continuously constructs local relations. Both views produce intermediate sub-expressions. These sub-expression representations are aligned using multi-granularity contrastive learning.} 
\label{Fig.approach} 
\end{figure*}

\indent We use the labeled mid-order equation to generate two different sequences $Y^{pre}\ \text{=}\ \{y^{f}_{1}, y^{f}_{2}, \cdots, y^{f}_{L}\}$ and $Y^{post}\ \text{=}\ \{y^{b}_{1}, y^{b}_{2}, \cdots, y^{b}_{L}\}$ using pre-order and post-order traversal. As shown in Figure \ref{Fig.f1}, we treat the $Y^{pre}$ as the label for the top-down process and the $Y^{post}$ is for the bottom-up process training.\\
\indent \textbf{Global shared Embedding} Firstly, we design three types of global shared embedding matrix: text word embedding $E^{w}$, quantity word embedding $E^{q}$, mathematical operator embedding $E^{op}$. Text embedding and quantity word embedding are extracted from the pre-trained language model \citep{devlin-etal-2019-bert, liu2019roberta}, and operator embeddings are randomly initialized. Besides, all constant word embeddings are also randomly initialized and added to $E^{q}$. As shown in Figure \ref{Fig.approach}, three global embeddings are shared by two reasoning processes. Then, text embeddings $E^{w}$ are fused into a target vector $t_{root}$ by the Bidirectional Gated Recurrent Unit (GRU) \citep{cho2014learning}, where $t_{root}$ means the global target for top-down reasoning. Quantity embeddings $E^q$ is for quantity relation construction in bottom-up reasoning.\\
\indent \textbf{Top-down view using Pre-order} The top-down view is a global-to-local decomposition that follows the pre-order equation $Y^{pre}$ (e.g., $-, \times, +, 2, 3, 4, 5$). This process is similar to \citet{xie2019goal}. Starting from the root node, each node needs to conduct \emph{node prediction}, and the operator node also conduct \emph{node decomposition}, e.g., in Figure \ref{Fig.f1}, root node predicts its node type is ``operator'' and output token is ``$-$'' and then is decomposed into two child nodes. Two child nodes are predicted to ``$\times$'' in step $2$ and ``5'' in step $7$.\\
\indent \emph{\textbf{Node prediction}} Each node has a target vector $t_{n}$ decomposed from their parent (for root node, $t_n\ \text{=}\ t_{root}$), and then calculates the node embedding $e_{n}$ and node output $y_{n}$ based on $t_{n}$ and global shared embedding $E^{w}$, $E^{op}$, $E^{q}$:
\begin{equation} \label{equ_s}
\begin{aligned}
e_{n} &= Attention(\text{MLP}^{e}(t_{n}), E^w)\\
s(op_i)\! &=\! \text{MLP}^{s}([e_{n};e^{op}_{i};e_{n}\!\circ\!e^{op}_{i}])\!,e^{op}_{i}\!\!\in\!\! E^{op}\\
s(q_j)\! &=\! \text{MLP}^{s}([e_{n};e^{q}_{j};e_{n}\!\circ\!e^{q}_{j}]),e^{q}_{j}\!\in\! E^{q} 
\end{aligned}
\end{equation}
where $;$ means the concatenation operation and $\circ$ denotes the element-wise product between two vectors. $\text{MLP}^{e}$ calculates the node embedding from target and $\text{MLP}^{s}$ computes the score of predicted output ($y_n\ \text{=}\ op_i$ or $q_j$) using the node embedding and the corresponding embedding ($e^{op}_{i}$ or $e^{q}_{j}$). $s(op_i)$ and $s(q_j)$ are the scores of the current node predicted to be $op_i$ and $q_j$.\\
\indent \emph{\textbf{Node decomposition}} After node prediction, any operator node ($y_n\ \text{=}\ op$) needs to be decomposed into two child nodes using their target vector $t_{n}$ and corresponding embedding $E^{op}[y_n]$:
\begin{eqnarray}
t_{n_l}, t_{n_r} = \text{MLP}^d([t_n;E^{op}[y_n]])
\end{eqnarray}
where $\text{MLP}^{d}$ is used for left and right child nodes decomposition, and $t_{n_l}$ and $t_{n_r}$ represent the target vectors of two child nodes.\\
\indent As shown in Figure \ref{Fig.approach}, the top-down process repeats above two steps: each node first predicts its own output, then the operator node is decomposed into two child nodes, and child nodes continue its node prediction. If any child node is still an operator node, the decomposition continues until the quantity node. The objective is to minimize the negative log-likelihood of training data $(T,Y)$ using the pre-order equation $Y^{pre}=\{y^f_1,y^f_2,\cdots,y^f_L\}$:
\begin{equation}
\begin{aligned}
L_{t2b} &= \!\sum_{T,Y}^{D} -log\  P\left(Y^{pre}\mid T\right)\\
        &= \!\sum_{T,Y}^{D} -log \prod_{n=1}^{L} P\left(y^f_n\mid E^{w,op,q},t_n\right)
\end{aligned}
\end{equation}
where $P(y_n^f\mid *)$ is the predicted probability of $y_n^f$ in node prediction, which is computed from all possible $s(op)$ and $s(q)$ (Equation \ref{equ_s}) by Softmax. The pre-order equation has $L$ tokens, so the top-down process also requires $L$ times of node prediction.\\ 
\indent \textbf{Bottom-up view using Post-order} 
The down-up view is a relation construction process that follows the post-order expressions $Y^{post}$ (e.g., $2,3,+,4,\times,5,-$). Inspired by \citet{jie2022learning}, we devise a concise bottom-up process. The sub-expression is treated as a relation mapping, i.e. operator is the math relation between two quantities, e.g.,$(2,3)\!\rightarrow \!+$. Thus, as shown in Figure \ref{Fig.approach}, in each iteration we map two quantities to a specific operator for a sub-expression, and then use this sub-expression as a new quantity for the next iteration, e.g., $(q_{(2,3,+)},4 \!\rightarrow \!\times)$. Specifically, in step $t$, a quantity pairs ($q_{i}$ and $q_{j}$) and a operator ($op_{k}$) form a relation mapping ($q_{i},q_{j}\rightarrow op_{k}$), we get their embeddings from the $E^{q}$ and $E^{op}$, i.e.,$e^{q}_{i},e^{q}_{j}\! \in\! E^{q}_{t}$ and $e^{op}_{k}\! \in\! E^{op}$, where $E^{q}_{t}$ is the embedding of the all quantity words at step $t$. We first fuse two quantity embeddings, and then with operator embedding:
\begin{equation}
\begin{aligned}
h^{j}_{i} &\!= \!\text{MLP}^{h}\left( [e^{q}_{i}; e^{q}_{j}; e^{q}_{i}\circ e^{q}_{j} ]\right)\!\\
\label{b2t-subexp}
e^{B2T}_{q_i,q_j,op_k} \!\!&= \!\text{MLP}^{m}\!\left([h^{j}_{i}; e^{op}_{k}; h^{j}_{i} \circ e^{op}_{k}]\right)\!
\end{aligned}
\end{equation}
where $e^{B2T}_{q_i,q_j,op_k}$ means the embedding of the sub-expression, $\text{MLP}^{h}$ fuses two quantity embeddings into $h^{j}_{i}$ and $\text{MLP}^{m}$ fuses $h^{j}_{i}$ with operator $op_k$.\\ 
\indent Then, to select the best mapping from all possible combinations of quantity pairs and operators, we score sub-expression based on its embedding:
\begin{equation}
\begin{aligned}
s\left(e^{B2T}_{q_i,q_j,op_k}\right) &= \text{MLP}^{r}\left(e^{B2T}_{q_i,q_j,op_k} \right)
\end{aligned}
\end{equation}
where $s(e^{B2T}_{q_i,q_j,op_k})$ means the score assigned to this sub-expression. Lastly, the selected sub-expression is added to $E^{q}_{t}$ and treated as a new quantity for next iteration, i.e. ,$E^{q}_{t+1}= E^{q}_{t} \cup \{e^{B2T}_{q_i,q_j,op_k}\}$.\\
\indent During training, we obtain the gold mapping ($y_{i}^{b},y_{j}^{b}\!\rightarrow \!y^{b}_{k}$) from the post-order equation $Y^{post}=\{y_1^b,y_2^b,\cdots,y_L^b\}$ and select the highest scoring mapping ($q_{i}^{m},q_{j}^{m}\!\rightarrow \!op^{m}_{k}$) from all combinations. The optimization is to maximize the score of the gold mapping in all possible combinations:
\begin{equation}
\begin{aligned}
L_{b2t}\!=\! \!\sum_{T,Y}^{D}\! \sum_{t=1}^{K}\! s\!\left(e^{B2T,t}_{q_{i}^{m},q_{j}^{m},op_k^{m}}\right)
\!- \!s\!\left(e^{B2T,t}_{y_{i}^b,y_{j}^b,y_{k}^b}\right)
\end{aligned}
\end{equation}
where $K$ denotes that equation $Y^{post}$ has $K$ times relation extraction in total.
\subsection{Consistent Contrastive Learning} \label{CCL}
The top-down reasoning provides a coarse-to-fine decomposition process in a flexible manner. In contrast, the bottom-up reasoning provides a local-to-global construction view step by step. Although the two views are reversed in process, they should be consistent regardless of the observation view. To this end, we use consistent contrastive learning to constrain the representations of the sub-expression generated in two independent views.\\ 
\indent \textbf{Multi-view Representation} For the top-down view, we fuse the embedding of the parent node with two child nodes in a sub-tree as a sub-expression representation. First, we calculate the parent node embedding $E^{op}[y_{p}]$, where $y_p$ means the predicted operator of the parent node. Then, the left child node embedding $e_{l}$ is calculated according to its node type. If the left child node is a quantity node, its embedding is $e_{l}\ \text{=}\ E^{q}[y_{l}]$, where $y_{l}$ means the predicted quantity. If left child is a operator node, the entire left subtree's representation is used as its embedding, i.e., $e_{l}\ \text{=}\ r^{T2B}_{l\text{=}sub\text{-}tree}$. The embedding of the right node $e_r$ is calculated in a similar way. Finally, we fuse three embeddings:
\begin{equation} \label{t2b_rep}
\begin{aligned}
r^{T2B}_{sub\text{-}exp} &=  \text{MLP}^{f}\!\left(\left[E^{op}[y_{p}\right];\!e_{l};e_{r}]\right)\!
\end{aligned}
\end{equation}
where $r^{T2B}_{sub\text{-}exp}$ means the sub-expression representation, and the entire sub-tree is treated as a new fusion node for the next calculation.\\
\indent For the bottom-up view, we directly use the embedding of the sub-expression obtained from each relation mapping (Equation \ref{b2t-subexp}) as its representation:
\begin{equation} \label{b2t_rep}
\begin{aligned}
r^{B2T}_{sub\text{-}exp} = e^{B2T}_{q_i,q_j,op_k}
\end{aligned}
\end{equation}
\indent \textbf{Multi-granularity Alignment} We align two views of the same sub-expression in multi-granularity. The sub-expression generated initially is the minimum granularity, and the maximum granularity is the complete equation representation. First, we select representations $r^{B2T}_{sub\text{-}exp}$ and $r^{T2B}_{sub\text{-}exp}$ from two views of the same sub-expression. Then, we project them into the same latent space ($h^{T}$ and $h^{B}$) and compute the similarity as consistent loss $L_{ccl}$. Finally, we repeatedly compute the consistent loss for each sub-expression:
\begin{equation} \label{align}
\begin{aligned}
h^{B}_{t} &= \text{MLP}^{c}(r^{B2T}_{sub\text{-}exp_t})\\
h^{T}_{t} &= \text{MLP}^{c'}(r^{T2B}_{sub\text{-}exp_t})\\
\mathcal{L}_{ccl}&=\!\sum^{D}_{T,Y}\! \left[\sum_{t=1}^{K}\! -\frac{\left \langle h^{T}_{t} ,h^{B}_{t} \right \rangle}{\left\| h^{B}_{t} \right\|_{2} \cdot \left\| h^{T}_{t} \right\|_{2}} \right]
\end{aligned}
\end{equation}
where $K$ denotes the total number of sub-expressions in the top-down view, and $<,>$ means dot product of two vectors for similarity. By alignment, two reasoning processes constrain each other at multiple granularities and jointly learn a more accurate and complete representation. We provide a detailed example (Figure \ref{Fig.appendix} in Appendix) to show the whole process.\\
\indent \textbf{Augmentation}
We argue that external math rules are essential for understanding diverse equations, e.g., different questions with the similar calculation logic are sometimes labeled as $(q_{1}+q_{2})\times q_{3}$ and sometimes as $q_{1}\times q_{3}+q_{2}\times q_{3}$. It is challenging to train with those labels. This inconsistency caused by diverse equations may impair performance. So we add a knowledge-enhancing augmentation (KE-Aug) process, which actively injects math laws for alleviating the impact of diversity.
Specifically, we exert deformations on all equations using a mathematical law, generating a new equation. Then, both new and origin samples are used for training, e.g., we use the multiplicative distributive law to convert all equations containing $(q_1\pm q_2)\times q_3$ into $q_1 \times q_3 \pm q_2\times q_3$. After that, the inconsistency is alleviated and the model can learn a similar representation for the equivalent equation.
\subsection{Training and Inference} 
During KE-Aug, we only use multiplicative distributive law as external knowledge for augmentation. Then, all samples are converted into the pre-order and post-order expressions. During training, to minimize the loss function $L = L_{t2b} + L_{b2t} + L_{ccl}$, we train three processes: top-down reasoning, bottom-up reasoning, and consistent contrastive learning simultaneously from scratch. During inference, we discard the bottom-up model and use top-down reasoning to compute the final prediction. Since top-down view is a generative model with more flexibility to generate diverse predictions than classification-based model (bottom-up) and also gain higher accuracy in our multi-view training framework (discussed in Section 4.2). 
\section{Experiments}
\begin{figure}[t] 
\centering 
\includegraphics[width=0.45\textwidth]{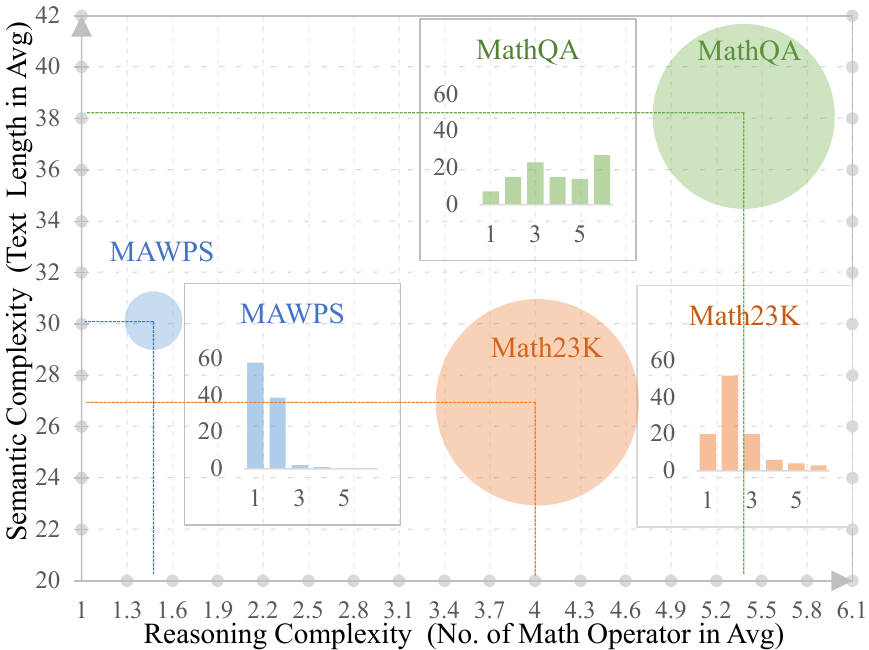} 
\caption{Statistics for the three datasets. The x-axis denotes the average number of math operators and the y-axis denotes the average text length. The area of the circle indicates the size of the dataset. Three histograms mean the distribution of the operators within the dataset.} 
\label{Fig.dataset} 
\end{figure}

\textbf{Datasets} We evaluate our method on three standard datasets across two languages: MAWPS \citep{koncel2016mawps}, Math23K \citep{wang2017deep}, and MathQA \citep{amini-etal-2019-mathqa}. Math23K and MathQA are two widely used large datasets that contain 23k Chinese mathematical problems and 20k English mathematical problems, respectively, and MAWPS only contains 1.9k English problems. We follow \citep{tan2021investigating,jie2022learning} to preprocess some unsolvable problems in the dataset. We consider five operators for the datasets: \emph{addition, subtraction, multiplication,  division, exponentiation} as previous works did.\\ 
\indent The number of mathematical operations is used to measure the reasoning complexity and the text length denotes the semantic complexity. In Figure \ref{Fig.dataset}, we plot the average reasoning complexity (x-axis) and semantic complexity (y-axis) of the three datasets in the two-dimensional plane. The MathQA is the hardest to solve as it has the highest semantic complexity and reasoning complexity. In contrast, the MAWPS is the easiest to answer, as almost all problems require only two mathematical operations. The Math23K dataset is the largest, with moderate reasoning complexity.\\
\indent \textbf{Baselines} We divide the baselines into the following categories: seq2seq, seq2structure, iterative relation-extraction (I-RE). Besides, we also consider the methods that use contrasting learning for generation (CL-Gen). In seq2seq, \citet{li-etal-2019-modeling} (\textbf{GroupAttn}) applied multi-head attention approach in seq2seq model. \citet{tan2021investigating} used multilingual BERT and an LSTM-based decoder (\textbf{mBERT}). \citet{lan2021mwptoolkit} utilized Transformer framework for MWP (\textbf{BERTGen, RoGen}). \citet{shen-etal-2021-generate-rank} proposed a multi-task approach (Gen-Rank). In seq2structure, \citet{xie2019goal} proposed seq-to-tree generation (\textbf{GTS}) method. \citet{zhang2020graph} (\textbf{Graph2Tree}) introduced GCN to encode numerical relations. \citet{patel-etal-2021-nlp,liang2021mwp} offered pre-trained language versions  (\textbf{Roberta-GTS, Roberta-G2T, BERT-Tree}). \citet{qin-etal-2021-neural} introduced a neural symbolic approach (\textbf{Symbol-Solver}) based on GTS. A hierarchical architecture extracts features for GTS \citep{yu-etal-2021-improving} (\textbf{H-Reasoner}). In I-RE, \citet{cao2021bottom} used DAG-structure to extract the relation (\textbf{DAG)}. \citet{jie2022learning} introduced a powerful RE framework to deduce relation. (\textbf{RE-Deduction}). In CL-Gen, \citet{ijcai2021-485} adopted a teacher model for discrimination (\textbf{T-Dis}). \citet{li2021seeking} proposed a prototype learning. (\textbf{CL-Prototype}).\\
\indent \textbf{Training Details} We adopt Roberta-base and Chinese-BERT from HuggingFace \citep{wolf-etal-2020-transformers} for multilingual datasets as previous works. We use an AdamW optimizer \citep{kingma2014adam,loshchilov2018decoupled} with a 2e-5 learning rate, batch size of 12, and beam search of size 4. All experiments were set up on an Nvidia RTX 3090 24GB. Following most previous works, we report the average answer accuracy (five random seeds) with standard deviation using the test set for Math23K and MathQA, and the 5-fold cross-validation performance on Math23K and MAWPS.
\subsection{Results}
\begin{table}[t]\small
\centering
\begin{tabular}{c|l|c|c}
\toprule[1pt]
                         &\textbf{Model}  &\textbf{ Acc.}   &\textbf{ 5-fold.}  \\ \midrule[0.5pt]
\multirow{5}*{\rotatebox{90}{Seq2Seq}}   &GroupAttn\citep{li-etal-2019-modeling}&69.5&66.9\\
                                         & mBERT*\citep{tan2021investigating} 	&75.1&-\\  
                                         & BERTGen*\citep{lan2021mwptoolkit}     &76.6&-\\  
                                         & RoGen* \citep{lan2021mwptoolkit}	&76.9&-\\   
                                  &Gen-Rank*$\diamondsuit$\citep{shen-etal-2021-generate-rank}	&85.4&-\\      
                                         \midrule[0.5pt]

\multirow{5}*{\rotatebox{90}{Structure-Gen}}    & GTS \citep{xie2019goal} &75.6&74.3\\
                                              & Graph2Tree\citep{zhang2020graph}	&77.4&75.5\\  
                                              &Symbol-Solver\citep{qin-etal-2021-neural} &-&75.7\\  
                                              & BERT-Tree*\citep{liang2021mwp} &84.4&82.3\\  
                                              & H-Reasoner*\citep{yu-etal-2021-improving} &83.9&82.2\\  \midrule[0.5pt]
\multirow{2}*{\rotatebox{90}{CL-G}}   & T-Dis*\citep{ijcai2021-485}  &79.1 &77.2\\
                     & CL-Prototype* \citep{li2021seeking}  &83.2&- \\ \midrule[0.5pt] 
\multirow{2}*{\rotatebox{90}{I-RE}}   & DAG* \citep{cao2021bottom} &77.5&75.1    \\
                     &RE-Deduction*\citep{jie2022learning}  &85.4&83.3 \\ \midrule[0.5pt]

\multirow{1}*{\rotatebox{0}{-}}   & Multi-view* (ours) &\makecell[c]{\textbf{87.1} \\ \scriptsize$\pm$ 0.29}  &\makecell[c]{\textbf{85.2} \\ \scriptsize $\pm$  0.38 }\\ 
\bottomrule[1pt]
\end{tabular}
\caption{Results on Math23k. $*$ means using pre-trained language model. $\diamondsuit$ means our reproduction.}
\label{tab:math23k}
\end{table}

\begin{table}[t]\small
\centering
\begin{tabular}{c|l|c}
\toprule[1pt]
                         &\textbf{Model}  &\textbf{ Acc.}     \\ \midrule[0.5pt]
\multirow{2}*{\rotatebox{0}{Seq2Seq}}   & GroupAttn$\diamondsuit$\citep{li-etal-2019-modeling}&70.4\\
                                         & mBERT*\citep{tan2021investigating} 	&77.1\\    \midrule[0.5pt]
\multirow{3}*{\makecell[c]{Structure \\ -Gen}}   & GTS$\diamondsuit$ \citep{xie2019goal} &71.3 \\
                                              & Graph2Tree$\diamondsuit$\citep{zhang2020graph}	&72.0\\  
                                              & BERT-Tree*\citep{liang2021mwp} &73.8\\    \midrule[0.5pt]

\multirow{1}*{CL-Gen}   & CL-Prototype* \citep{li2021seeking}  &76.3 \\ \midrule[0.5pt]
\multirow{1}*{I-RE}   & RE-Deduction*\citep{jie2022learning}  &78.6 \\ \midrule[0.5pt]
\multirow{1}*{\rotatebox{0}{-}}   &      Multi-view*(ours)  &     \makecell[c]{\textbf{80.6} \\ \scriptsize$\pm$ 0.17 }\\ 
\bottomrule[1pt]
\end{tabular}
\caption{Test accuracy comparison on MathQA.}
\label{tab:mathqa}
\end{table}

\begin{table}[htbp]\small
\centering
\begin{tabular}{c|l|c}
\toprule[1pt]
                         &\textbf{Model}  &\textbf{ Acc.}     \\ \midrule[0.5pt]
\multirow{5}*{\rotatebox{0}{Seq2Seq}}   &GroupAttn\citep{li-etal-2019-modeling}&76.1\\
&Gen-Rank\citep{shen-etal-2021-generate-rank}&84.0\\
                                         & Transformer\citep{vaswani2017attention}	&85.6\\  
                                         & BERTGen*\citep{lan2021mwptoolkit}     &86.9\\  
                                         & RoGen*\citep{lan2021mwptoolkit}	&88.4\\   \midrule[0.5pt]

\multirow{5}*{\makecell[c]{Structure \\ -Gen}}    & GTS \citep{xie2019goal} &82.6 \\
                                              & Graph2Tree\citep{zhang2020graph}	&85.6\\  
                                              & Roberta-GTS*\citep{liang2021mwp} &88.5\\  
                                              & Roberta-G2T*\citep{liang2021mwp} &88.7\\  
                                              & H-Reasoner*\citep{yu-etal-2021-improving} &89.8\\  \midrule[0.5pt]

\multirow{1}*{CL-Gen}   & T-Dis*\citep{ijcai2021-485}  &84.2 \\ \midrule[0.5pt]
\multirow{1}*{I-RE}   & RE-Deduction*\citep{jie2022learning}  &92.2 \\ \midrule[0.5pt]
\multirow{1}*{\rotatebox{0}{-}}   & Multi-view* (ours)   &    \makecell[c]{\textbf{92.3} \\ \scriptsize $\pm$0.16 } \\ 
\bottomrule[1pt]
\end{tabular}
\caption{5-fold cross-validation results on MAWPS.}
\label{tab:tabel_mawps}
\end{table}

\indent  As shown in Table \ref{tab:math23k}, \ref{tab:mathqa}, we observe our method achieves consistent improvements over the strong baselines across multiple datasets, with +1.7\% improvements on Math23K, +1.9\% improvements on 5-fold Math23K, +2.0\% improvements on MathQA. The improvement is particularly significant when our method is evaluated on larger and more complex datasets, like MathQA, which includes many GRE problems requiring complex reasoning. We achieve the greatest improvement on this most difficult dataset. It demonstrates the reliable reasoning ability of our method. Additionally, although the MAWPS dataset is small and simple, we still obtain a slight boost (+0.1\%) compared to the other baselines in Table \ref{tab:tabel_mawps}.\\
\indent Compared with three single-view methods: seq2seq, seq2structure and I-RE, our method is more stable and outperforms all of them. Although, the I-RE method performs the best among all single-view methods, it still lags behind ours by almost 2\% (RE-deduction) on average. In addition, the performance of the other two single-view methods is unstable: on the simpler but larger dataset Math23K, seq2structure achieves comparable accuracy with seq2seq, but lags behind ours by 2.7\% (BERT-Tree), 1.7\% (Gen-Rank), respectively. In contrast, on the more complex dataset MathQA, seq2seq is better than seq2structure, but worse than ours by 3.5\% (mBERT*) and 6.8\% (BERT-Tree).\\
\indent Furthermore, we also observe that the method which adopts contrastive learning (CL-Prototype) is considerably lower than ours by 3.9\% (Math23K) and 4.3\% (MathQA). It suggests that our multi-view design is pretty effective for math reasoning, and contrastive learning can play a more significant role in our consistent multi-view framework. A fine-grained analysis can be found in Section \ref{ana exp}.
\subsection{Ablation Experiments}
Through the above experiments, we found that data augmentation can alleviate inconsistency between different instances and multi-view contrastive learning can alleviate inconsistency between different views of an instance. To better illustrate the contribution of each module, we devise several variant models and evaluate them on Math23K.

As the Table \ref{ablation_table} shows, Multi-view means that the model contains both top-down and bottom-up reasoning processes, and keeps both views consistent through global shared embedding and contrastive learning. KE-Aug means we adopt 
equation augmentation for training. (1) Our proposed method (Multi-view and KE-Aug) achieves an accuracy of 87.1\% in top-down view. In contrast, the bottom-up view does not show any performance gains. (2) We remove the multi-view alignment, and two reasoning views are completely independent and both are trained using augmented data. The performance of the top-down view dropped by 1.7\% (87.1\% to 85.4\%), while bottom-up view performance also dropped by 1.2\% (85.2\% to 84\%). (3) In contrast, we remove the data augmentation module in that the two reasoning views can learn more precise representations by a consistent contrastive learning. In this case, there is a slight decrease in the top-down view (-0.6\%), while the accuracy of the bottom-up is instead improved by +1.1\%. (4) Moreover, after removing KE-aug and Multi-view, it only consists of two completely independent reasoning processes and can only be trained on the original inconsistent dataset. The two views achieve 84.9\% and 85.1\% accuracy respectively, which are comparable to the other single-view baselines. 

These ablation experiment clearly reveal that data augmentation brings small or negative improvement on single-view approaches, but multi-view alignment can maximize the effect of augmentation. We suspect that it may be because the bottom-up view focuses more on local features, and the data augmentation brings multiple local relations, thus making such local features more difficult to extract. Therefore, during training, we use consistent contrastive learning and data augmentation to train multi-view processes. As for the inference process, we directly use the top-down view as the final prediction model.
\begin{table}[t]\small
\renewcommand\arraystretch{1.15}
\centering
\setlength\tabcolsep{2pt}
\begin{tabular}{l c c}
\toprule
     \textbf{Variant} & \textbf{Top-down} & \textbf{bottom-up}\\
\midrule[0.25pt]
                         Multi-view\ \& KE-Aug            &\textbf{87.1} &85.2       \\ 
w/o Multi-view                      &85.4   &84  \\
w/o KE-Aug                      &86.5   &86.3  \\
w/o KE-Aug \& Multi-view   &84.9   &85.1 \\

\bottomrule
\end{tabular}
\caption{Ablation study on Math23K.}
\label{ablation_table}
\end{table}
\subsection{Analysis Experiments}\label{ana exp}
\begin{figure}[t] 
\centering 
\includegraphics[width=0.5\textwidth]{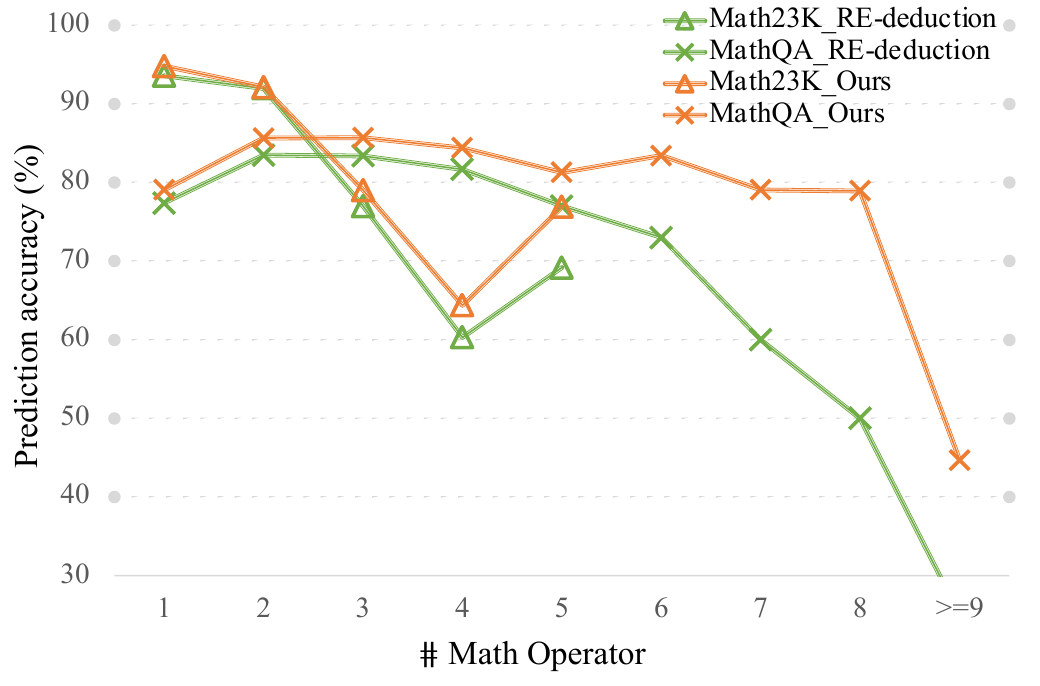} 
\caption{A fine-grained analysis of model performance on subsets with different reasoning complexity.} 
\label{Fig.fine-grain} 
\end{figure}

\begin{figure}[t] 
\centering 
\includegraphics[width=0.48\textwidth]{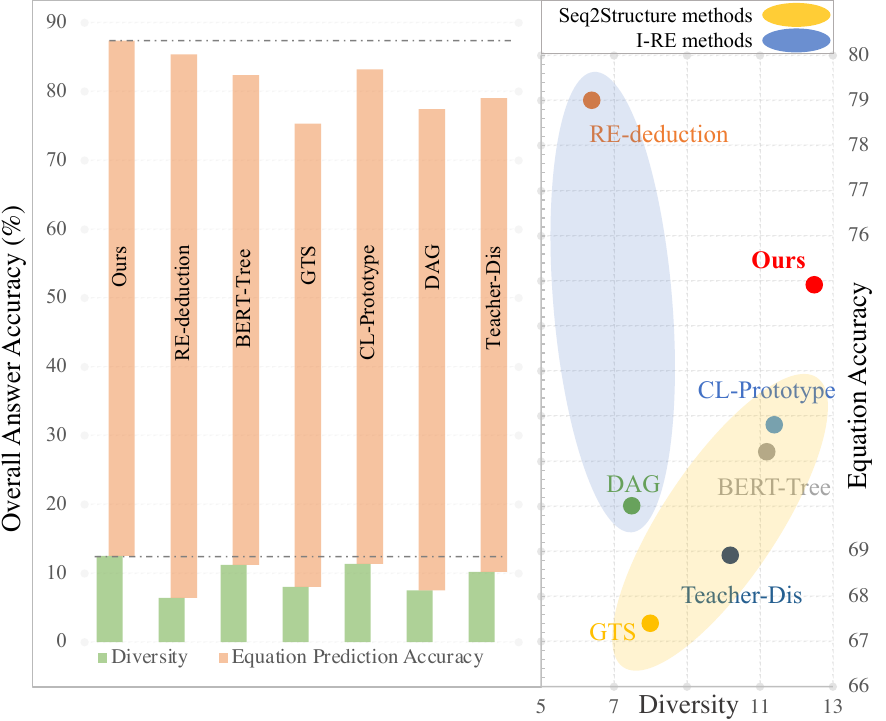} 
\caption{We split the overall accuracy into two parts, i.e., \emph{Equation accuracy} and \emph{Diversity}. Our overall accuracy and diversity are the highest (Left). The seven methods are plotted in a 2D plane according to two metrics (Right). Our method performs well on both metrics.} 
\label{Fig.Attribution} 
\end{figure}

\textbf{Fine-grained Comparison} To verify that our method can handle more complex math problems, we conduct a fine-grained comparison with the best baseline (RE-deduction) on two challenging datasets (MathQA and Math23K). Specifically, we calculate the performance of the subset divided by the number of mathematical operators.\\ 
\indent As shown in Figure \ref{Fig.fine-grain}, our proposed method gains consistent improvements over the baseline across all subsets. In particular, on the more complex MathQA, we still maintain high prediction accuracy ($\geq 78\%$) over hard problems (number of operators $\geq 4$ ), but the performance of the baseline drops dramatically, e.g., on the most complex subsets with 8 and 9 operators, our performance outperforms the baseline by nearly 20\% and 28\%. A similar trend can be observed on math23K, i.e., our method achieves more significant results on more difficult subset, with 2.06\% improvements on the 3 operators subset, 4.08\% on the 4 operators subset and 7.7\% on the 5 operators subset. The superiority we achieve on these difficult samples demonstrates strong global generation and accurate local mapping capabilities for math reasoning.\\ 
\indent \textbf{Performance Attribution Analysis} To further demonstrate our method can achieve a high prediction accuracy while also predicting equations with diversity, we split the overall precision into two parts: equation precision and diversity. Equation precision indicates the proportion of samples in the test set whose prediction is exactly the same as the label equation. Contrary to this, diversity counts those samples whose prediction are different from the label, but also derive the correct answer, e.g., $Y^{pred}=\{+,-,\times,2,4,5,\times,3,4\}$, $Y^{label}=\{-,\times,+,2,3,4,5\}$.\\ 
\indent As Figure \ref{Fig.Attribution} shows, the overall precision (87.1\%) and diversity (12.4\%) of ours both are the highest among the seven methods, and equation precision is only inferior to RE-deduction. Besides, we plot the equation precision (x-axis) and diversity (y-axis) on a two-dimensional plane. We find that the I-RE methods (RE-deduction and DAG) has low diversity but relatively high equation precision. In contrast, the seq2structure methods (GTS, Teacher-Dis, BERT-Tree) generate more diverse results with low equations precision. However, our method performs well in both diversity and equation precision.\\ 
\indent We also provide some examples in the case study (Figure \ref{case} in Appendix). This experiment illustrates that each of these single-view approaches has specific limitations, either lacking fine-grained mapping or global diverse generation capabilities. Our multi-view approach can incorporate the merits of both views, achieving precise and versatile solving. 
\section{Conclusion}
We treat the pre-order traversal of math equation as a top-down view and the post-order equation as bottom-up view. Two reasoning views are naturally existing and both are necessary for complex mathematical reasoning. We design a multi-view reasoning containing a top-down decomposition and a bottom-up construction and ensure the consistency of the two views through contrastive learning. This consistent multi-view 
design can endow us with a complete and precise semantics-to-equation mapping. Experiments on standard datasets show that our framework achieves new state-of-the-art performance, especially demonstrating reliable generation capabilities on long and complex problems. 

\section*{Limitations}
There are two main limitations of our work: first, although we design two reasoning processes during training: top-down and bottom-up, we discard the bottom-up process when inferring and only adopt the prediction from the top-down reasoning. In future work, we will explore how to select the best prediction from both views. Second, our multi-view reasoning process is capable of generating more diverse equivalent equations, but this generation process is not controllable, and it is not clear for now what underlying factors control different generation patterns.

\section*{Acknowledgments}
This work is supported by the Key Research and Development Program of Zhejiang Province, China (No. 2021C01013), the National Key Research and Development Project of China (No. 2018AAA0101900), CKCEST, and MOE Engineering Research Center of Digital Library.

\bibliography{custom}
\bibliographystyle{acl_natbib}
\appendix
\clearpage
\renewcommand\thefigure{\Alph{section}\arabic{figure}}    
\setcounter{figure}{0}    
\section{Appendix}
\label{sec:appendix}
\subsection{Details for Consistent Contrastive Learning}
\begin{figure}[t]
\begin{center}
\includegraphics[width = 0.45\textwidth]{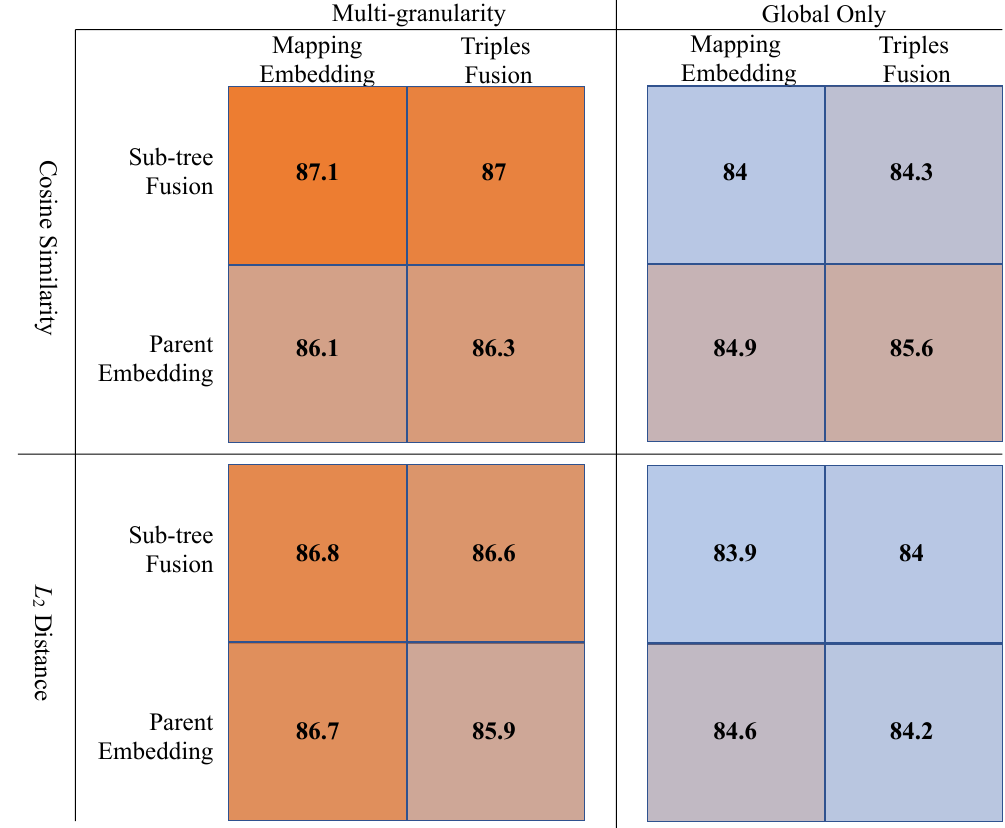}
\end{center}
\caption{Evaluation on Math23K using multiple configurations of consistent contrastive learning.}
\label{detail}
\end{figure}

We investigate the design of consistent contrastive learning for multi-view alignment. As shown in Figure \ref{detail}, we evaluate four factors on Math23K:\\ 
\indent \textbf{Metric for alignment}. We consider two metrics for alignment: cosine similarity and $L_2$ distance. The former is a simplification of the conventional contrastive metric with only positive instances.\\
\indent \textbf{Granularity of alignment}. As shown in Equation \ref{align}, we use multi-granularity sub-expressions for alignment. Besides, we also show the results of aligning two views only using the global equation representation.\\
\indent \textbf{Top-down representation}. We investigate how to obtain the representation of sub-expressions. We design two types of representation for the top-down view. First, as shown in Equation \ref{t2b_rep}, we use sub-tree fusion to get the representation for each sub-expression, which is denoted as \emph{sub-tree fusion}. Besides, we treat the embedding of the parent node $e_n$ (Equation \ref{equ_s}) as a representation for this sub-expression. We denote it as \emph{parent embedding}.\\ 
\indent \textbf{Bottom-up representation}. For the bottom-up process, there are also two options for its representations. As shown in Equation \ref{b2t_rep}, we use the embedding of the relation mapping as the representation, which denotes as \emph{mapping embedding}. Besides, we fuse the concatenation of the three embeddings using $\text{MLP}$ layer: $r^{B2T}_{sub\text{-}exp}= \text{MLP}([e_i^q;e_j^q;op_k^q])$, which denotes it as \emph{triples fusion}. 
\subsection{Visualization}
\begin{figure*}[hb]
\begin{center}
\includegraphics[width=1\textwidth]{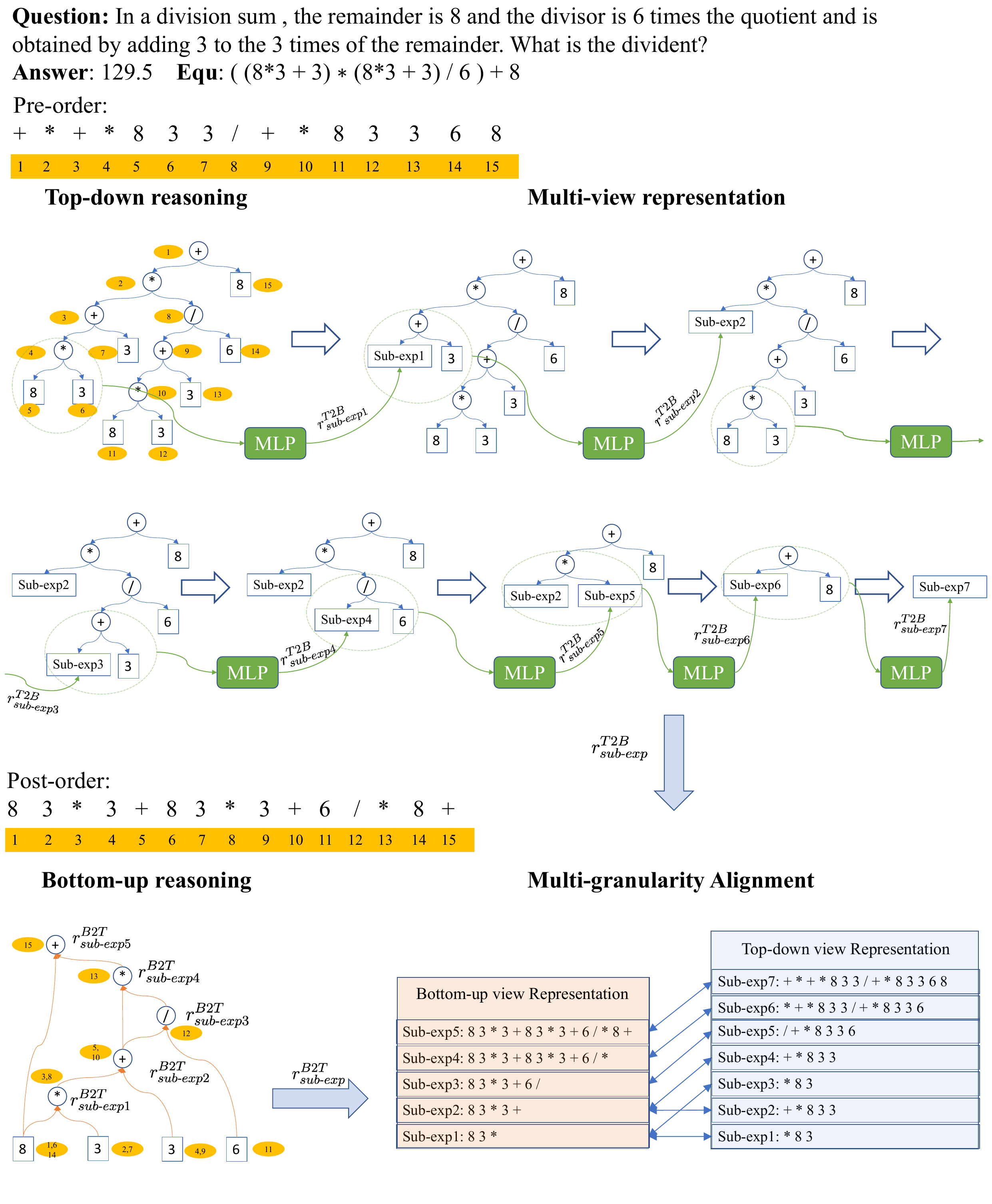}
\end{center}
\caption{A MathQA example of multi-view reasoning and consistent contrastive learning process. It contains independent reasoning processes of two views, the computations of sub-expressions representation, and multi-granularity alignment.}
\label{Fig.appendix} 
\end{figure*}

In Figure \ref{Fig.appendix}, we show an example from MathQA. The top-down process breaks down the overall problem through 15 reasoning procedures which are exactly the same as the pre-order traversal. Each reasoning step includes two steps:  node prediction and node decomposition, until the leaf node (quantity nodes). Meanwhile, the bottom-up view predicts the entire equation after five relation extractions following the post-order equation. Two reasoning views work in reverse order.\\ 
\indent Then, in the consistent contrastive learning process, the top-down view continuously computes the sub-expression representation based on the sub-tree fusion. For bottom-up reasoning, we directly use the embeddings from each relation extraction as representations. Since the bottom-up process reuses the previously constructed sub-expressions (step $8$ and $10$), it generates fewer sub-expressions than the top-down process. Finally, all sub-expressions representations from both views are aligned in the same latent space.\\
\subsection{Case Study} 
\begin{figure*}[hb]
\begin{center}
\includegraphics[width = 0.95\textwidth]{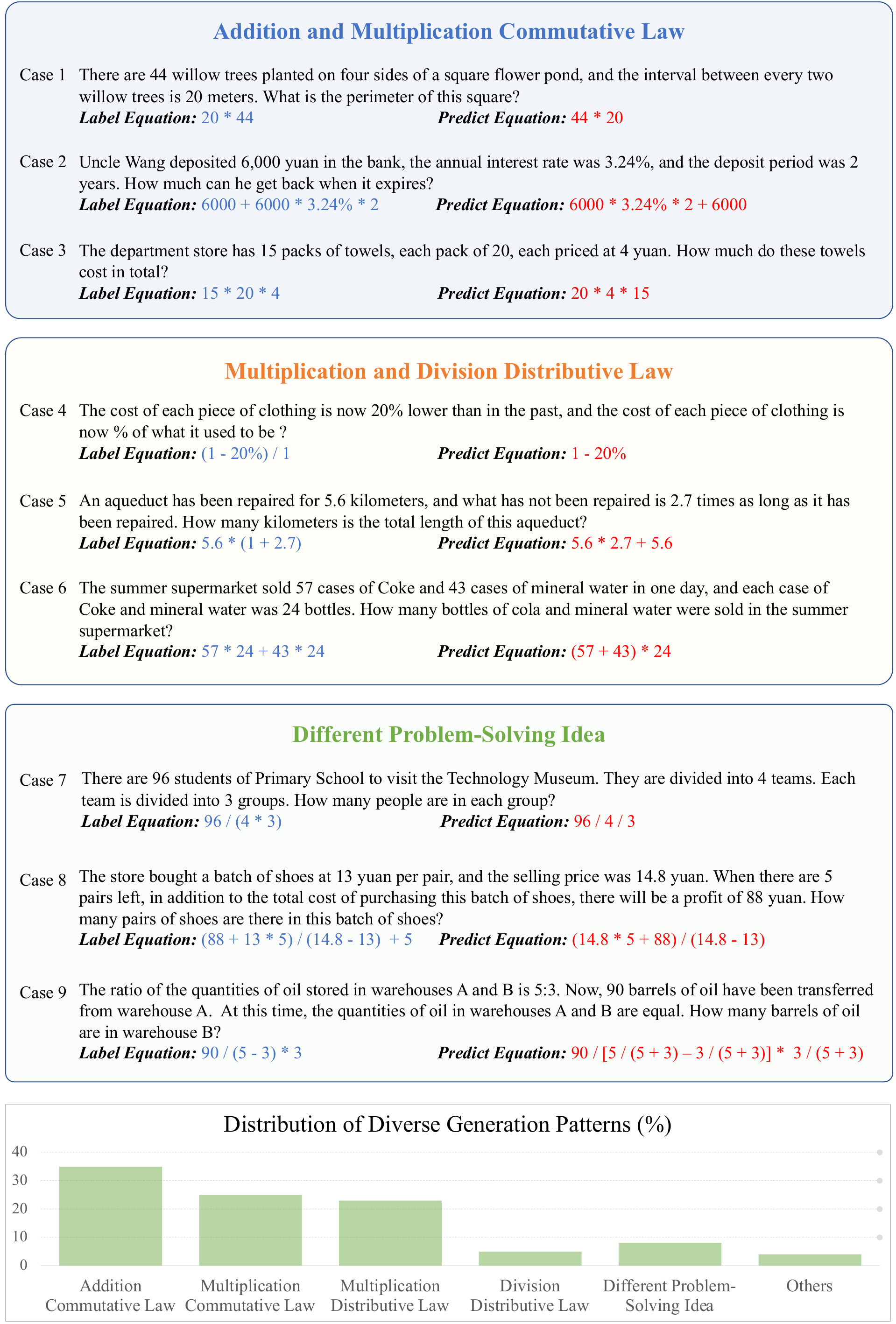}
\end{center}
\caption{Nine examples demonstrate the capability of our approach for generating equivalent but non-labeled equations. At the bottom, we count the distribution of the six generation patterns among all equivalent equations. Each pattern represents a mathematical deformation using a specific mathematical law. This diverse generation indicates that our model can understands the underlying mathematical relation and generates reasonable equation based on mathematical laws.}
\label{case}
\end{figure*}

We perform a case study to demonstrate the capability of generating diverse equations. As Figure \ref{case} shows, the algorithm generates equivalent equations that are not the same as the labeled equations. 
Most of these predicted equivalent equations can be derived from labeled equations by simple mathematical deformations, e.g., $(57+43)\times24$ and $57\times24+43\times24$ in case 6. In addition to simple deformations, our algorithm also can solve complex problems using the different solving ideas, e.g., in case 8, it starts from a simpler reasoning idea and solves the problem correctly.\\ 
\indent At the bottom of Figure \ref{case}, we also count the deformation pattern distributions among all diverse prediction equations. We summarize six patterns: \emph{addition and multiplication commutative law}, \emph{multiplication and division distributive law}, \emph{different problem-solving idea} and \emph{others}. Then we manually identify the deformation patterns of each equivalent equation predicted by ours. We discover that more than half of the equivalent equations ($\geq60\% $) can be derived from additive or multiplicative commutative law deformations. Nearly 30\% of the equivalent equations can be derived by deforming the distributive law and about 8\% belong to the different solving ideas (e.g., cases 8 and 9). It shows that our multi-view method has mathematical reasoning capabilities and can be applied to solve complex mathematical problems. 
\end{document}